\documentclass{article}

\usepackage{microtype}
\usepackage{graphicx}
\usepackage{subfigure}
\usepackage{booktabs} 

\usepackage{hyperref}

\usepackage[accepted]{icml2021}

\usepackage{comment}

\icmltitlerunning{ARRID: ANN-based Rotordynamics for Robust and Integrated Design}

\begin{document}

\twocolumn[
\icmltitle{ARRID: ANN-based Rotordynamics for Robust and Integrated Design}



\icmlsetsymbol{equal}{*}

\begin{icmlauthorlist}
\icmlauthor{Soheyl Massoudi}{to}
\icmlauthor{Jürg Schiffmann}{to}
\end{icmlauthorlist}

\icmlaffiliation{to}{Ecole Polytechnique Fédérale de Lausanne (EPFL), Laboratory for Applied Mechanical
Design, CH-1015 Lausanne, Switzerland}

\icmlcorrespondingauthor{Soheyl Massoudi}{soheyl.massoudi@epfl.ch}

\icmlkeywords{Machine Learning, ICML}

\vskip 0.3in
]



\printAffiliationsAndNotice{}  

\begin{abstract}
The purpose of this study is to introduce ANN-based software for the fast evaluation of rotordynamics in the context of robust and integrated design. It is based on a surrogate model made of ensembles of artificial neural networks running in a Bokeh web application. The use of a surrogate model has sped up the computation by three orders of magnitude compared to the current models. ARRID offers fast performance information, including the effect of manufacturing deviations. As such, it helps the designer to make optimal design choices early in the design process. The designer can manipulate the parameters of the design and the operating conditions to obtain performance information in a matter of seconds. 
\end{abstract}

\section{Introduction}
The design of gas-bearings supported turbomachinery is a complex task, since it involves the design of different components such as the gas-bearings, the rotor, the electric motor, and the turbomachinery aerodynamics. In the early design process, multi-objective optimization can be used to find a Pareto optimal design \cite{massoudi2022robust}. However, before sending the design to production, the engineers often need to perform small modifications to accommodate manufacturing processes, which may impact the performance of the system. To be able to immediately assess the impact of such modifications, the ARRID framework was developed with the following goals in mind:
\begin{enumerate}
    \item Fast and accurate computation through the use of ensembles of artificial neural networks
    \item Pertinent visual information on the performance robustness of the turbocompressor
    \item Acceleration of the engineer's rework process to speed up the launch to production
\end{enumerate}

\section{Integrated and robust design}
Design in engineering is complex. It involves sequential mappings from the technical scope of specifications to the process variables via the defined functional requirements and matching design parameters \cite{nordlund2015axiomatic}. Different performance metrics and objectives need to be assessed under constraints. Ideally, all the subsystems need to be designed together to obtain an optimal design, framed integrated design \cite{schiffmann2015integrated}. Often, designing for the nominal point is not sufficient as operating conditions and manufacturing are subject to tolerances. Hence, there is a need for robust design against manufacturing deviations, which should satisfy the constraints of the technical scope of specifications, while maintaining excellent performance \cite{hasenkamp_review_2009}.\par
Design is also an iterative task. For the design of turbomachinery as an example, the designer starts from a 0D model to obtain a first guess of a subset of the system design dimensions \cite{mounier_data-driven_2018}. From there, the design is refined along with sequential steps with models of increased fidelity (2D, 3D). Should the design not perform as intended, the designer must make modifications earlier in the design sequence and run new calculations. Obtaining a correct initialization is therefore paramount in making the design process effective. 

\section{Gas-bearings supported turbocompressors}
Target applications for small-scale turbomachinery are compressors for fuel cells and heat pumps. Herringbone grooved journal bearings (HGJB) are laser-engraved gas-bearings allowing the supported rotor shaft to spin at high speeds, while achieving a high lifetime with minimal maintenance \cite{gu2020review}. The gas-bearings are self-lubricated by the thin fluid layer surrounding the rotor and entrapped by the stator. Hence, there is no need for oil lubrication, eliminating the risk of pollution in the process. \par
A limitation of the application is the sensibility of the rotordynamic system stability to manufacturing deviations \cite{multiObjRobManuTol}. The rotor system is in the tens of millimeter range, whereas the laser engraved grooves are in the order of micrometers. While laser engraving is accurate to the micrometer on the groove depth, meeting nominal bearing clearances in the order of 5-10 micrometers requires significant manufacturing effort. In this context, a robust design is achieved by maximizing the feasible region of the gas-bearings supported turbocompressor under deviations of local bearing clearance and groove depth. \par

A finite difference scheme is used to compute the bearings stiffness and damping of the gas-bearings using small perturbations from the concentric position. Excitation frequencies are swept and the logarithmic decrement, metric of stability, is computed via an intersection method \cite{geertseindhoven}. Although this process is fast - in the order of magnitude of the second - it is not fast enough to provide immediate information about the robustness of the design. As a consequence, we use feed-forward artificial neural networks (ANNs) to build a global surrogate model to accelerate the computations, while retaining reasonable accuracy \cite{Goodfellow-et-al-2016}.

\section{Structure of the surrogate model}
The surrogate model is made of ensembles of six ANNs blocks \cite{ganaie2021ensemble}. The ANNs are trained using a data-driven approach using a model of a rigid gas-bearings supported rotor \cite{schiffmann2010effect}. A genetic algorithm is used to identify the hyperparameters of the network to minimize the training loss function \cite{harvey_lets_2017, papavasileiou_systematic_2021}. Four ensemble blocks are used to predict the stability of the rotor. A first classifier predicts if the system is in an excited mode and a second one predicts whether the design is stable. A regressor then predicts the dimensionless excitation frequency known as the whirl speed ratio. The last regressor predicts the logarithmic decrement. Each of the four excitation modes is treated by a sequence of four ensemble ANNs blocks.

\section{An interactive visualization of the performance of the system using bokeh}
\begin{figure*}[ht]%
\centering
\includegraphics[width=\linewidth]{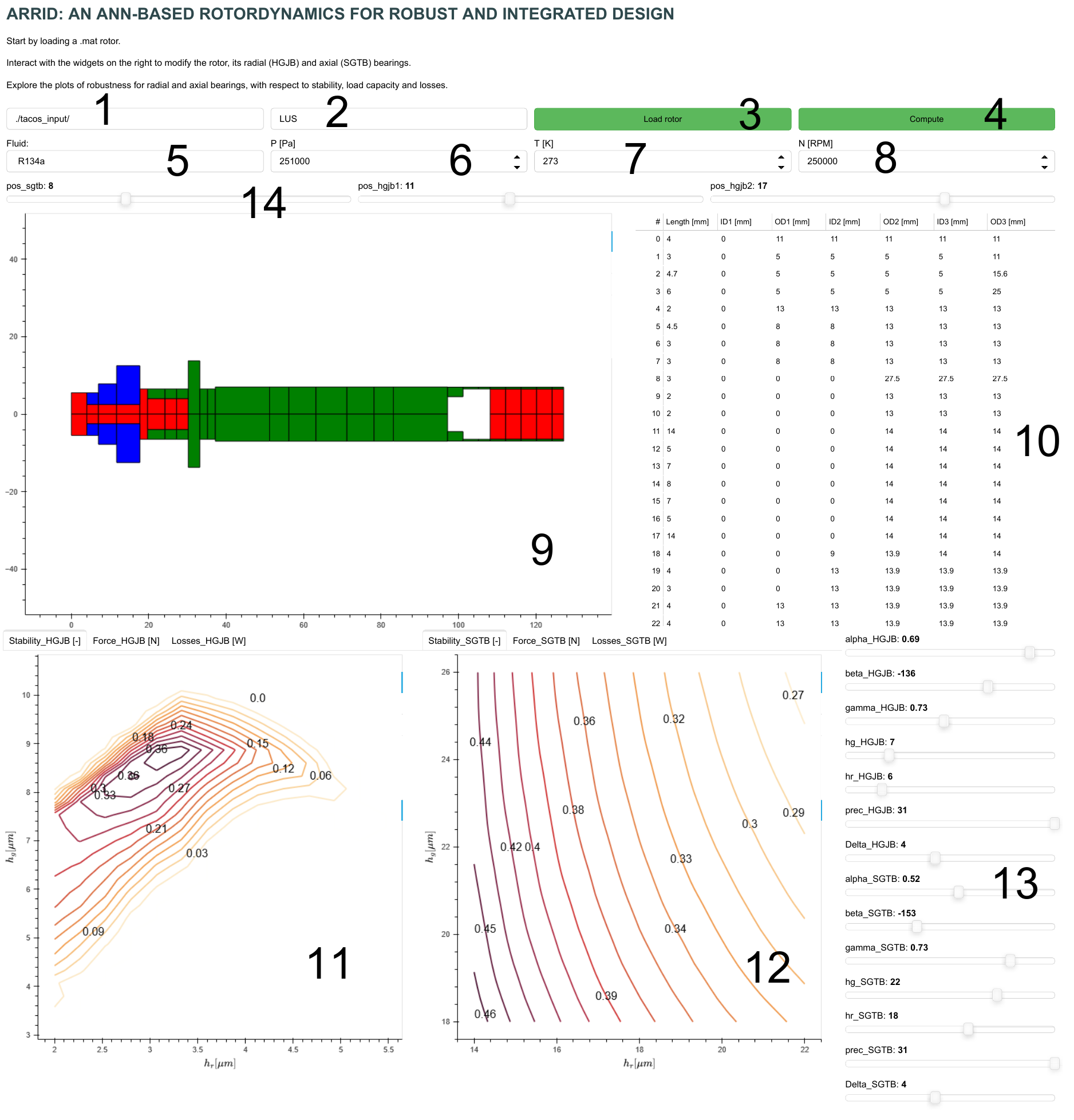}
\caption{ARRID is written in Python using the Bokeh library. It is displayed in the local browser upon execution. The user selects the folder (1) where the rotor defined by its filename (2) is located. It is loaded using the Button widget (3). Operating conditions are defined in (5), (6), (7) and (8). The rotor and its dimensions in mm are displayed on the interactive figure (9). Its geometry can be further analyzed and edited in the DataTable widget (10). Journal bearings (HGJB) and axial bearing (SGTB) geometries and computation accuracy are selected with sliders (13). Sliders widgets are also used to define the position of the bearings (14). Pressing the "Compute" button (4) starts the evaluation of the performance and the robustness of the rotor against manufacturing deviations. The results are displayed on interactive contour plots for the journal bearings (11) and the axial bearing (12). Tabs allow the user to switch between the performance metrics of stability, load capacity and power losses.}
\label{fig:ARRID_soft}
\end{figure*}

ARRID is a Bokeh dashboard where the user can load, edit, and visualize a rotor and its performance in stability, load capacity, and power losses. Bokeh is a Python library used to create interactive visualizations in web browsers \cite{bokeh-2022} and is used here as a web application running on a server. ARRID features are described on Figure \ref{fig:ARRID_soft}.

The user can load a rotor design by typing the name of the file to be loaded in the Text widget and pressing the Load Button widget. The rotor is formatted as a series of cylindrical elements with up to three layers of different materials embedded in a Matlab structure. It is then converted to a Python dictionary. Each cylindrical element has the following properties which can be changed by selecting and modifying the DataTable widget :

\begin{itemize}
    \item $L$: length in $m$
    \item $d_{i=1,2,3}$: inner diameter of layer $i$ in $m$
    \item $D_{i=1,2,3}$: outer diameter of layer $i$ in $m$

\end{itemize}

Two of the cylinders can then be defined as the journal bearings and a third one as the thrust bearing. The following parameters are updated respectively for both journals (same topology), and the axial bearing using the Sliders widgets:

\begin{itemize}
    \item $\alpha$, the groove-ridge width ratio
    \item $\beta$, the groove angle
    \item $\gamma$, the grooved region ratio
    \item $h_g$, the groove depth in $\mu mu$
    \item $h_r$, the local bearing clearance in $\mu m$
    \item $\Delta h_r$, the deviation from $h_r$ in $\mu m$
    \item $\Delta h_g$, the deviation from $h_g$ in $\mu m$
\end{itemize}

The operating conditions can also be defined by the user via the Select and Spinner widgets:

\begin{itemize}
    \item $fluid$, the fluid used to lubricate the bearings 
    \item $Pa$, the ambient pressure at the bearings in $Pa$
    \item $T$, the ambient temperature at the bearings in $K$
    \item $N$, the target rotational speed in $m$
\end{itemize}

Callbacks are implemented so that output plots are updated as the computation button is pressed in the dashboard:

\begin{enumerate}
    \item A plot made of rectangles representing the rotor
    \item A tab for the plots of the robustness of the radial bearings
    \item A tab for the plots of the robustness of the axial bearing
\end{enumerate}

In this call, the Python dictionary describing the rotor is updated. Calls are made to the functions computing the mass and moments of inertia of the whole rotor. A sensitivity analysis for local bearing clearance $h_r$ and local groove depth $h_r$ is performed as well as a sweep of the nominal speed. From the dimensional values, dimensionless groups are computed and inputted into the ANNs. The ANNs compute the whirl speed ratio and logarithmic decrement for the whole input dataset. This data is then fed to the plots of the robustness of stability, load capacity, and losses against manufacturing deviations for radial and axial bearings respectively. The plot describing the rotor is also updated. The whole process takes less than a minute when using a modern GPU to perform the computations against dozens of minutes when using the base models. 

\section{Conclusion}
This framework helps the designer achieve on-the-fly modifications to the design and evaluate its overall nominal and robust performance.
\begin{enumerate}
    \item The use of surrogate models accelerates by two orders of magnitude the computation of the performance of the gas-bearings supported turbocompressor
    \item The dashboard offers pertinent visual information to help the user understand the performance of the gas-bearing supported turbocompressor upon variation of its design parameters and operating conditions
    \item The designer is offered an interactive tool to help in decision making by validating at early stage modifications needed for manufacturing
\end{enumerate}

\nocite{langley00}

\bibliography{main.bib}
\bibliographystyle{icml2021}

\end{document}